\documentclass[conference]{IEEEtran}
\IEEEoverridecommandlockouts
\usepackage{cite}
\usepackage{amsmath,amssymb,amsfonts}
\usepackage{algorithmic}
\usepackage{graphicx}
\usepackage{textcomp}
\usepackage{xcolor}

\usepackage{booktabs}
\usepackage{makecell}
\usepackage{float}

\def\BibTeX{{\rm B\kern-.05em{\sc i\kern-.025em b}\kern-.08em
    T\kern-.1667em\lower.7ex\hbox{E}\kern-.125emX}}
\begin{document}

\title{Real-Time Mask Detection Based on SSD-MobileNetV2\\
}

\author{\IEEEauthorblockN{1\textsuperscript{st} Chen Cheng}
\IEEEauthorblockA{\textit{School of Computer Science} \\
\textit{China University of Geosciences}\\
Wuhan, China \\
chengchen@cheney.cc}
}

\maketitle

\begin{abstract}
After the outbreak of COVID-19, mask detection, as the most convenient and effective means of prevention, plays a crucial role in epidemic prevention and control. An excellent automatic real-time mask detection system can reduce a lot of work pressure for relevant staff. However, by analyzing the existing mask detection approaches, we find that they are mostly resource-intensive and do not achieve a good balance between speed and accuracy. And there is no perfect face mask dataset at present. In this paper, we propose a new architecture for mask detection. Our system uses SSD as the mask locator and classifier, and further replaces VGG-16 with MobileNetV2 to extract the features of the image and reduce a lot of parameters. Therefore, our system can be deployed on embedded devices. Transfer learning methods are used to transfer pre-trained models from other domains to our model. Data enhancement methods in our system such as MixUp effectively prevent overfitting. It also effectively reduces the dependence on large-scale datasets. By doing experiments in practical scenarios, the results demonstrate that our system performed well in real-time mask detection.
\end{abstract}


\section{Introduction}
Over the past two years, COVID-19 has swept the globe, and it is relevant to everyone. Many studies\cite{howard2020face} have shown that the virus can be transmitted through saliva in the air during sneezing \cite{galbadage2020does}. Currently, because no effective treatment has been developed, prevention is the only way to control the spread of the disease. There are many ways to prevent outbreaks, but wearing masks \cite{eikenberry2020mask} is considered the simplest and most effective way to prevent COVID-19. And in \cite{verma2020visualizing}, it is indicated that proper wearing of masks can effectively mitigate the spread of the virus. Wearing a mask not only helps to prevent the spread of the virus, but also protects healthy people from possible airborne infections. So governments are emphasizing that people wear masks in public places. However, there are still some people who do not  follow wearing masks in public place because of the trouble and discomfort. In order to enforce everyone to wear a mask in the public place such as large supermarkets, hospitals, it is necessary to check the mask-wearing status by hand.

However, manual detection of mask-wearing is time-consuming. The automatic mask detection system can maintain stable operation for a long time and cost little price. According to the large amount of data obtained by the automated mask detection system, it is possible to establish a machine learning model for detection mask wearing. 

Inspired by the successful application of object detection technology in many practical scenarios, it is easy to think of applying it to the scene of mask detection. Object detection can locate and detect a certain type of object in the target image. For example, many places that require identity verification, such as stations, are often equipped with automated face detection devices, which greatly improves the efficiency of detection. Thus, the technology of object detection can also be used to deal with the mask detection problem in face images. Automatic mask detection can be achieved with the help of object detection algorithms. An efficient mask detection system is particularly important, but at this stage, there is a lack of efficient automatic real-time mask detection devices. In the following article, we will analyze the existing mask detection models.

As a popular research direction in the field of object detection, after analyzing a large amount of related work on mask detection, this paper finds that mask detection faces many challenges, which mainly include the following aspects: 1) There are a huge diversity of masks caused by different colors, shapes etc. Besides, there are many kinds, and different people wear different standards. Moreover, the mask blocks part of the facial features, reducing the feature points that can be extracted on the face. 
2) Real-time mask detection requires high accuracy and speed. The accuracy of identification is an important guarantee for epidemic prevention and control, and wrong judgment may lead to a new round of epidemics. Since the automatic mask detection device is generally set up in public places with high traffic, slow detection speed may lead to crowded people. 
3) Lack of high-performance computing equipment. The real-time mask detection system often runs in embedded devices with scarce hardware resources, which poses a huge challenge to the efficiency of the model. The system must be able to continuously and stably operate in a low physical configuration environment. 
4) There are a few datasets that can be used for training. At this stage, the number of correctly labeled face mask datasets is relatively small, and most of them need to be obtained and processed manually. In summary, it is challenging to achieve high accuracy and fast speed detection by a small number of datasets on devices with poor hardware resources. Apart from that, detecting diverse masks poses a huge challenge to researchers. 

Most of the previous studies have focused on identifying the identity of the person wearing the mask \cite{chen2018face}. Many researchers have proposed a series of methods from different aspects. Mohamed et al. make use of the speech recognition method to judge the audio information of the speaker\cite{mohamed2022face}, which is less invasive for mask recognition.
Qin et al. combined the SR (Super-Resolution) network with the classification network for the first time, and proposed a mask-wearing state detection algorithm named SRCNet \cite{qin2020identifying}, This model can classify the detection results into three categories, namely Correct Wear (CFW), No Wear (NFW) and Incorrect Wear (IFW). Similarly, Jiang et al. proposed a dataset with three-class labels (PWMFD) \cite{jiang2021real}, and proposed the SE-Yolov3 model based on this dataset, replacing the mean squared error (MSE) with GIoU loss, and the feature extraction capability of the backbone network is enhanced, so that the efficiency and accuracy of model detection can reach a certain balance. However, these previous studies do not focus on judging whether the test object wears a mask, especially detecting people who do not wear a mask from the crowd. 

Even though mask detection has developed rapidly, compared with traditional computer vision fields such as face detection, there is still a lot of room for improvement in current mask detection. After analyzing the above-mentioned existing mask detection models, it is found that most of them have the following shortcomings: most of the current models focus on the accuracy of detection while ignoring the speed and the occupation of hardware resources; The detection speed is not fast enough, and there is a risk of people passing quickly in front of the machine but being ignored because the machine is not refreshed; The hardware resource consumption is large, traditional identification methods often build complex networks and require a lot of computation, and embedded mask identification devices are often difficult to support.

In order to overcome the shortage we mentioned above, we leverage SSD-MobileNetV2 and propose a new platform to solve these problems and detect the masks in real-time. When it comes to the lack of wealthy data, we use transform learning, which can migrate  some other object detection models to the field of mask detection. And then it is helpful to speed up the training progress and improve the accuracy. To prevent overfitting and to maximize the use of the limited dataset, we perform data augmentation methods on the dataset. In order to reduce the size of the neural network model, we use MobileNetV2 to replace the backbone network VGG-16 of SSD, so as to obtain a lighter network model. We list our contributions as follows:
\begin{itemize}
    \item We applied SSD-based MobileNet-V2 to the field of mask detection and establish a machine learning model for it.
    \item We developed a real-time mask detection system. The system can be able to detect mask-wearing status in all kinds of scenarios.
    \item We design the experiments and obtain results that can be satisfied with the practical mask-wearing tasks.
\end{itemize}

The rest of the article will unfold in the following structure: In Section 2, we will introduce a series of studies related to mask detection. In Section 3, we present the detailed structure of our model and how we build this model. Section 4 includes the acquisition of the dataset and data augmentation methods. The construction and evaluation of the system are also described in this section. Finally, Section 5 concludes what we have proposed in this article.

\section{Related Work}
\subsection{Object Detection}
With the development of deep convolutional neural networks \cite{xiao2020review}, object detection has gradually transitted from canonical methods to deep learning methods. Most of the current object detection technology is based on DCNN (Dynamic Convolution Neural Network), and capturing the position and classification of objects are its essence. With the continuous improvement of hardware computing power and the increasing number of excellent datasets, DCNN-based object detection has flourished. First, AlexNet\cite{krizhevsky2012imagenet} was proposed by Krizhevsky et al. and won the ILSVRC-2012 competition. This is the first time DCNN appeared in history. Then Girshick et al. proposed the RCNN(Regions with CNN) model\cite{girshick2014rich} based on this model, marking that DCNN has entered a new stage in the field of target detection. The object detection algorithm based on deep learning has ushered in rapid development, new network models are constantly being proposed \cite{tan2020efficientdet, kong2020foveabox}, and parameters such as recognition accuracy are continuously refreshed.

Generally, object detection is divided into two categories, namely one-stage object detection and two-stage object detection. Two-stage object detection consists of two steps: the first step is to locate the target and find the position of the target to be detected in the image or each frame of the video. The second step is to classify objects and classify the located objects according to their various characteristic processes. The representative algorithm of Two-stage is the aforementioned RCNN series, including the original RCNN model and other improved models such as Fast RCNN \cite{girshick2015fast} and Faster RCNN \cite{ren2015faster}. Although the accuracy of these algorithms can reach a relatively high level, the overall efficiency is relatively low due to the need for two processing. In order to solve this problem, Joseph et al. proposed Yolo (You Only Look Once), which is a model based on a single network \cite{redmon2016you}, and the one-stage algorithm officially entered the stage of history. This type of method omits the step of generating candidate regions, and directly detects the final result in a single time, so the speed is very fast, but the accuracy of positioning is sacrificed. Subsequently, the Yolo family continued to expand, and gradually developed YoloV2 and YoloV3, etc., and the accuracy and speed of detection were continuously improved. Although the speed of the Yolo series is greatly improved than that of RCNN, the accuracy rate is not high enough compared to the RCNN series. So in order to solve this problem, Liu et al. proposed SSD (Single Shot MultiBox Detector) \cite{liu2016ssd}, which borrowed some ideas from both Faster R-CNN and Yolo, and is more efficient than Yolo in terms of dense multi-targets and large objects. It is effective and can achieve high accuracy while meeting the speed requirements of real-time detection.

MobileNetV2 is an image classifier that provides lightweight and accurate predictions. In MobileNetV1, a different method from traditional convolution is depthwise separable convolution \cite{howard2017mobilenets}, which can greatly reduce the parameters of the model. MobileNetV2 adds Linear Bottleneck and Inverted Residual to the v1 version. As one of the core networks of target detection, MobileNet is often used to reduce the size of the model due to its lightweight characteristics. For example, Huang R et al. utilize MobileNet to mitigate YoloV3 \cite{huang2019rapid}, thus achieving a balance between speed and accuracy.

With the continuous development of object detection, it is also gradually applied in many important fields, such as: face detection, pedestrian detection \cite{zhao2017pedestrian}, automatic driving \cite{dai2019hybridnet}, medical decision-making \cite{jaeger2020retina}, video surveillance \cite{mhalla2018embedded}, etc.

\subsection{Mask Detection}
After the outbreak of Covid-19, masks have become an indispensable part of people's daily life, and mask detection has gradually become a popular research direction in computer vision\cite{sethi2021face}. In general, mask detection still belongs to the category of object detection. 
In recent years, many scholars have done a series of studies in the field of mask research. Loey et al. adopted a hybrid model \cite{loey2021hybrid}, in which one part uses the popular ResNet-50 in deep transfer learning for feature extraction, and the other part uses traditional machine learning algorithms such as SVM, decision tree, and ensemble learning. A feature extraction model with an end-to-end structure is constructed. High detection accuracy is achieved, but time consumption is not considered too much. Nagrath et al.
obtain better performance of classifying mask images by establishing the SSDMNV2 model leveraging a deep convolutional neural network.
~\cite{nagrath2021ssdmnv2}. Apart from that, the contribution of a good dataset to the model cannot be ignored. In view of the lack of the current mask dataset, Wang et al. \cite{wang2020masked} constructed three face datasets with masks, namely: MFDD, RMFRD, and SMFRD dataset. And a new algorithm was proposed, which made the face detection with masks reach 95$\%$ accuracy. Jignesh Chowdary et al. obtained very high accuracy on the Simulated Masked Face Dataset (SNFD) dataset by transfer learning from a pre-trained InceptionV3 model \cite{jignesh2020face}.

In general, the above models have improved mask detection in some aspects, and our method will draw on their experience and focus on real-time mask detection.

\begin{figure}[htbp]
    \centerline{\includegraphics[scale=0.2]{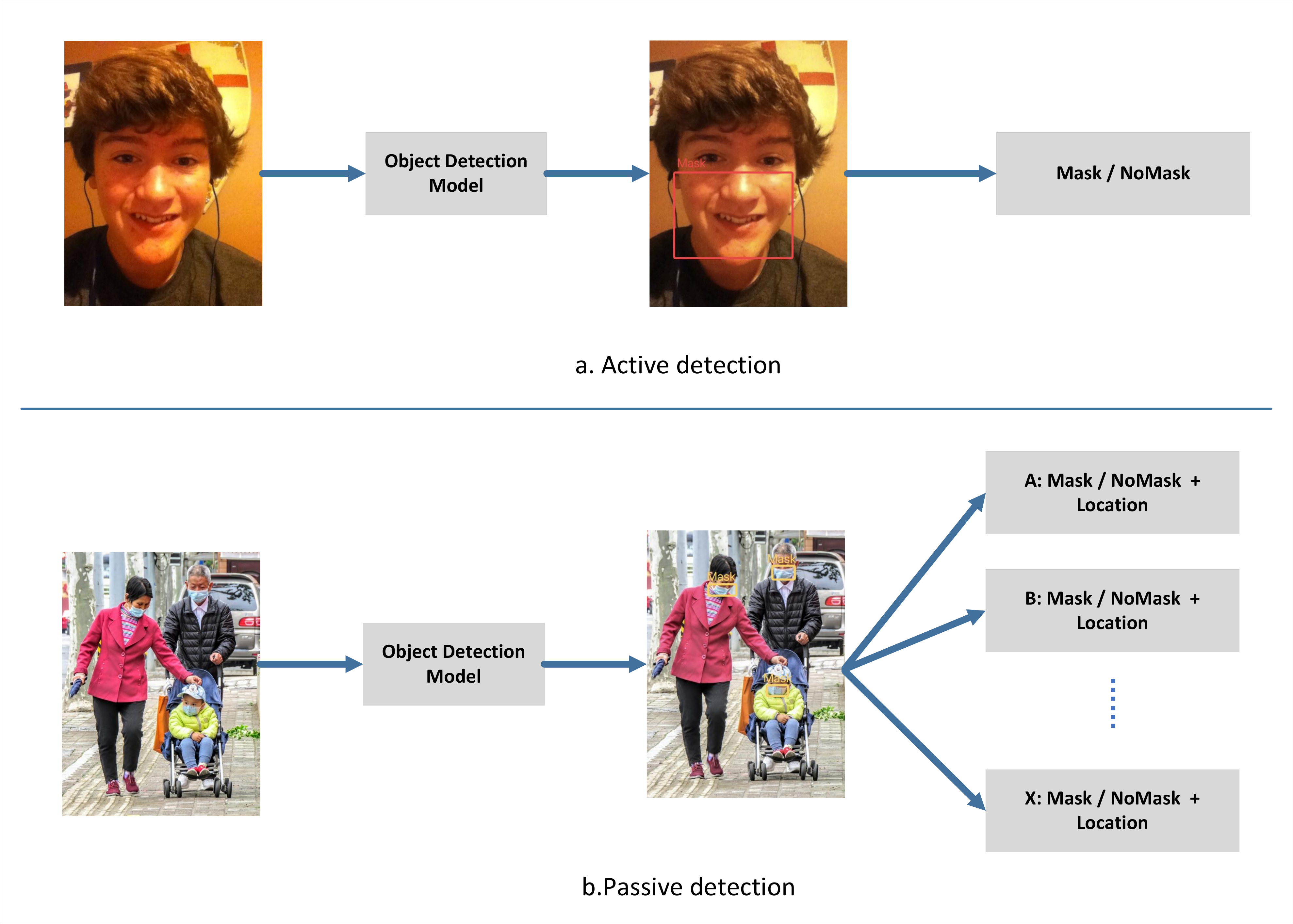}}
    \caption{Active and Passive Detection.}
    \label{active-passive-detection}
\end{figure}

Mask detection has always been considered a challenging task, and real-time mask detection brings us even greater challenges. As shown in Figure \ref{active-passive-detection}, real-time mask detection is generally divided into two types: active detection and passive detection. The former is mainly for large-scale detection in places with high traffic and high density. Each image used for processing contains multiple face-mask objects and often accounts for a small proportion. While active detection has only one target and collects images at close range, similar to the scene of face detection, only one result is returned per image. This paper mainly introduces more challenging passive detection techniques which are suitable for applications in public places.

\section{Mask Detection based on SSD-MobileNetV2}
Based on the analysis of a lot of relevant research, we find that most of the existing algorithms are designed to improve accuracy, and often ignore the hardware dependence brought by complex models. However, real-time mask detection is often installed in some embedded devices, which usually do not have high-end GPUs and cannot support a large number of complex calculations, so a lightweight network that can achieve high accuracy must be adopted.

\begin{figure}[htbp]
    \centerline{\includegraphics[scale=0.15]{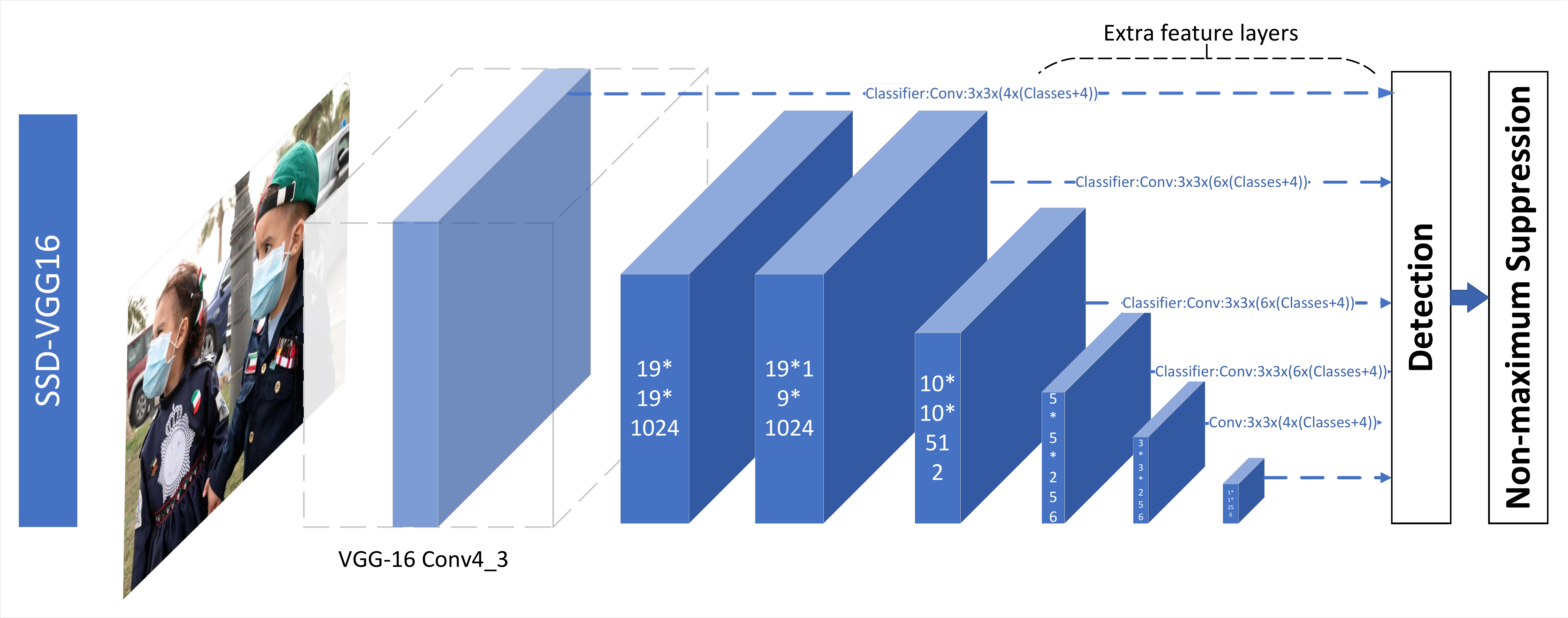}}
    \caption{SSD-VGG16}
    \label{ssd-vgg16}
\end{figure}

SSD (Single Shot MultiBox Detector) appeared after the Faster-RCNN and YoloV1 algorithms, inheriting many of the advantages of the two, balancing their performance in terms of speed and accuracy. SSD adds convolutional layers to get more features for detection. The VGG-16 network is the backbone of SSD. The VGG-16 network has a total of 16 layers. The first half is the superposition of the convolutional layers, the latter is the fully connected layer, and the last is the SoftMax layer for normalization. Similar to other One-stage models, the first step of the SSD model is to extract feature images using convolution operations. In order to extract feature images on each scale, SSD uses the last few layers of convolution to perform feature extraction according to the candidate frame obtained by the anchor. The target type and position are judged based on the candidate frame obtained by the anchor. The overall structure of the SSD is shown in figure:\ref{ssd-vgg16}

Although the effect of SSD has been relatively balanced, it is still difficult to run on low-configured embedded devices. The main reason is that the backbone network VGG-16 contains too many parameters and requires a lot of storage and running space to complete the detection task, which makes the entire model too large. So we consider replacing it with a more lightweight network.
In order to transfer deep learning-based object detection models from large servers to small mobile terminals, some lightweight models are proposed. Howard et al. first proposed MobileNetV1, which divides the object detection network structure into depth-wise convolution and point-wise convolution. 
%,as shown in Figure \ref{mobilenet-convolution} 
% \begin{figure}[htbp]
%     \centerline{\includegraphics[scale=0.3]{figure/mobilenet-convolution.pdf}}
%     \caption{Comparison between different convolutions}
%     \label{mobilenet-convolution}
% \end{figure}

The depth-wise convolution adopted by MobileNet has great advantages in the number of parameters and calculation speed. Although it reduces the amount of computation and parameters of the model, the model still continues the straight-up and straight-down characteristics of the VGG network. To this end, Sandler et al. proposed MobileNetv2 \cite{sandler2018mobilenetv2} based on MobileNetv1, which reduces the number of parameters by 20$\%$ and further improves the accuracy. Mobile Net v2 draws on the residual connection idea of ResNet and proposes a reverse residual structure based on this, which allows the network to allow smaller input and output dimensions and make the network deeper. It is worth noting that when the step size is 2, the series structure is directly adopted, and only when the step size is 1, there is a residual connection, as shown in Figure \ref{mobilenet-stride}.

\begin{figure}[htbp]
    \centerline{\includegraphics[scale=0.4]{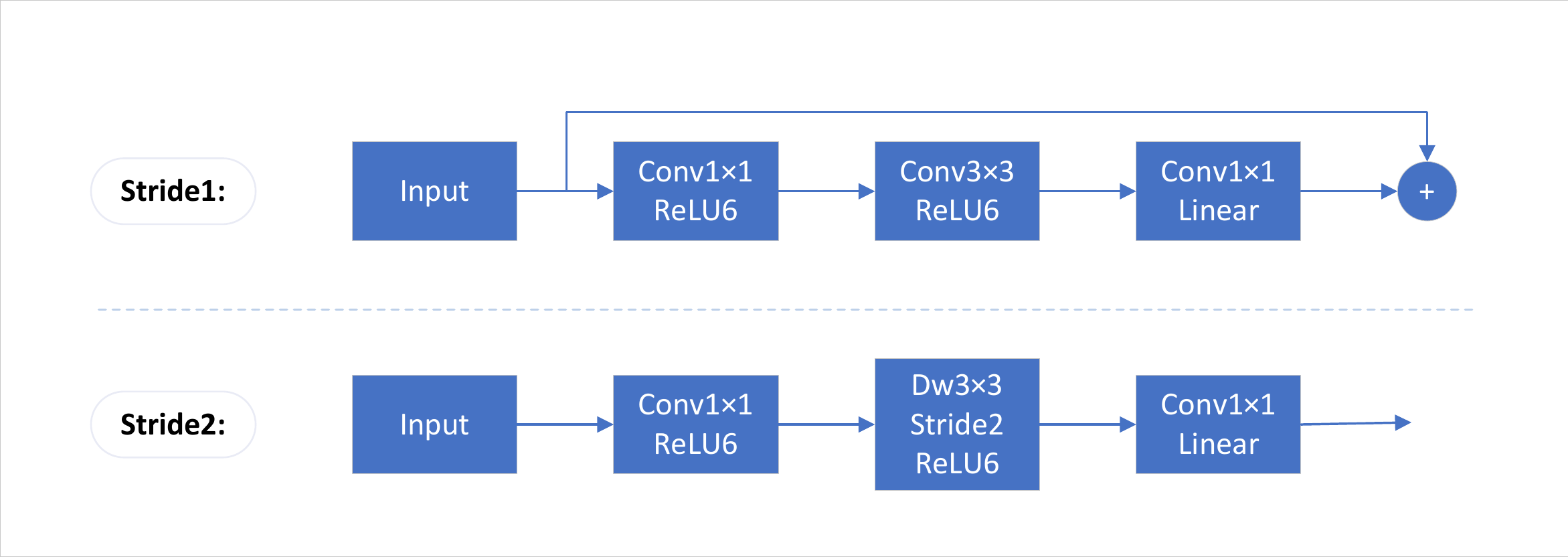}}
    \caption{mobilenet-stride}
    \label{MobileNetV2}
\end{figure}

As previously analyzed, the RCNN series can not meet the needs of real-time mask detection due to its large model and slow training. In contrast, the Yolo network is not only fast, but also has low resource consumption, so it is widely used in embedded devices. However, due to the special nature of our mask detection in epidemic prevention and control, it is difficult for Yolo to meet our needs for accuracy. Therefore, SSD, which can achieve a certain balance between accuracy and model size, has naturally become our first choice.

\begin{figure}[htbp]
    \centerline{\includegraphics[scale=0.15]{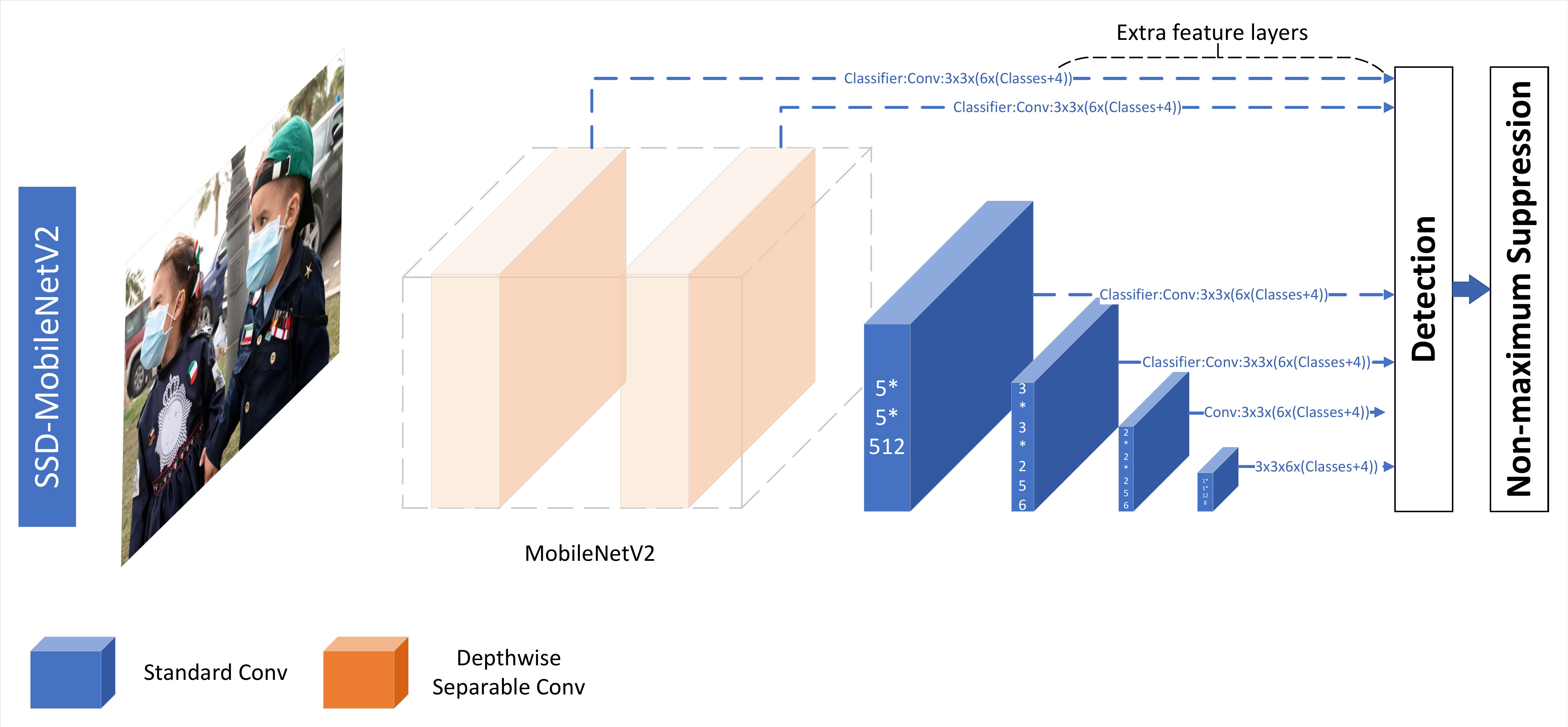}}
    \caption{SSD-MobileNetV2}
    \label{ssd-mobilenetv2}
\end{figure}

In order to further optimize the size of the network, we finally replace VGG-16 with MobileNetV2 as the backbone network of SSD, and finally get SSD-MobileNetV2 network as our model. The schematic diagram is shown in Figure \ref{ssd-mobilenetv2}

Since mask detection has just attracted extensive research in recent years, the existing face-mask datasets are not rich enough to support a large amount of training. In other fields of computer vision, such as face detection, there are already mature models and a large number of excellent datasets. Therefore, the idea of transfer learning is adopted to transfer existing models from other fields to mask detection, thereby enhancing the accuracy of the training model and reducing the dependence on datasets to a certain extent. At the same time, some data augmentation methods are used to effectively prevent the risk of overfitting and make more efficient use of limited data sets.

Overall, compared to other studies, the advantages of our work can be listed as follow. Firstly, we leverage SSD, which is much faster and more accurate, especially faster than Rcnn and more accurate than Yolo. Secondly, the backbone network MobileNetV2 we use is lightweight and can meet the requirements of embedded devices. Thirdly, transfer learning can improve performance and reduce dependence on large datasets. Last but not least, the system can detect masks in real time under complex scenarios.

\section{Experiment}
\subsection{Environment}
The real-time mask detection model proposed in this paper is based on the Python language. We set up the development environment and completed all the work of model training, validation as well as testing  on the Windows PC.

We list our experiment environment in table \ref{ExperimentEnvrionment}.
\begin{table}[hbp]
 \centering
 \caption{Experiment Environment}
 \label{ExperimentEnvrionment}
 \begin{tabular}{cp{5cm} lp{6cm}}
  \toprule
  Environment & Version \\
  \midrule
  Operation System & Windows 11 Professional x64 21H2 \\
  CPU & Intel(R) Core(TM) i7-9750H @2.60GHz \\
  RAM & 16 GB DDR4 \\
  GPU & NVIDIA GeForce RTX 2060 (6G)\\
  Development Tool & Pycharm Professional Edition 2022\\
  CUDA ToolKit & 11.2\\
  CuDNN & 8.1.0\\
  Tensorflow & 2.5.0\\
  OpenCV & 4.0.1\\

  \bottomrule
 \end{tabular}
\end{table}

\subsection{Data Set}
We get the dataset from Kaggle. The first part is "Face Mask Detection ~12K Images Dataset".\footnote{https://www.kaggle.com/datasets/ashishjangra27/face-mask-12k-images-dataset } This dataset includes nearly 12K images, and the dataset divides the images into different folders according to whether people wear masks or not. All images with masks are crawled from Google, and other images without masks are preprocessed from Jessica Liby Jessica Li. However, the proportion of faces in the entire dataset is too high, so the Face Mask Detection Dataset \footnote{https://www.kaggle.com/datasets/wobotintelligence/face-mask-detection-dataset}  is used. This dataset contains a large number of face mask images in complex scenes, and the proportion of faces is small. After simple filtering of the dataset, we obtained the original dataset of the system. This dataset consists of about 10,000 face mask pictures, 50$\%$ of which are faces with masks and 50$\%$ without masks. Each picture contains a variable amount of faces with or without a mask.

First of all: According to the Vicinal Risk Minimization (VRM) principle, in order to prevent overfitting, and also to maximize the use of limited data sets, we have adopted some methods of Data Augmentation. In addition to traditional techniques (eg: translation, rotation and mirroring), we also adopted MixUp method \cite{zhang2017mixup}.

\begin{figure}[htbp]
    \centerline{\includegraphics[scale=0.35]{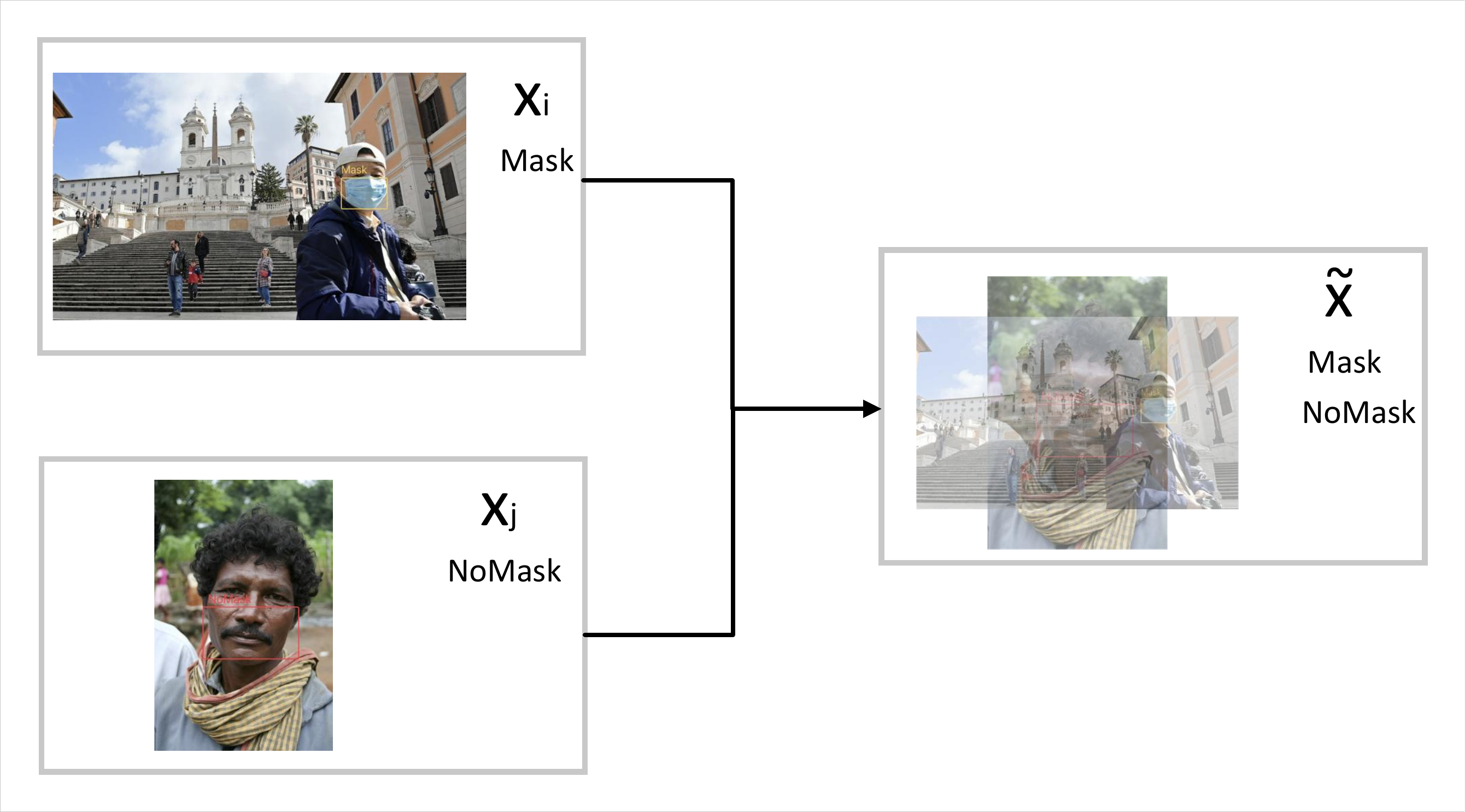}}
    \caption{Data Augmentation Method}
    \label{mixup}
\end{figure}

Mixup is a simple Data Augmentation method that is not related to the dataset itself. It can effectively improve the generalization ability of the model while being simple enough and not bringing too much additional burden to the model. In our system, the two sample-label data pairs to be processed are fused together(as shown in Figure \ref{mixup}) according to a certain ratio, and then a new image is obtained by linear interpolation, as shown in  Equation (\ref{mixup-xi}) and  Equation (\ref{mixup-yi}):

\begin{equation}
    \label{mixup-xi}
    \widetilde x = \lambda x_i + (1-\lambda)x_j
\end{equation}

\begin{equation}
    \label{mixup-yi}
    \widetilde y = \lambda y_i + (1-\lambda)y_j
\end{equation}
Where $x_i$,$x_j$ are raw input vectors, and $y_i$,$y_j$ are one-hot label encodings. ($x_i$,$y_i$) and ($x_j$,$y_j$) are two samples randomly selected from the training data, and $\widetilde x$ is a new sample after mixing $x_i$ and $x_j$. $\lambda \in [0,1]$ is the probability value, and $\lambda \sim Beta(\alpha,\alpha)$ obeys the Beta distribution whose parameters are all $\alpha$. In our experiments, we set the value of $\alpha$ to 0.4.

LabelImage is an image annotator that can be downloaded from GitHub and is easy to install. Markers are saved as XML files in PASCAL VOC format. The data set just obtained has no labels. We use the LabelImg tool \footnote{https://github.com/heartexlabs/labelImg} to label each image with Mask or NoMask, and use a rectangular box to mark the location of the mask. Make each picture have an XML file to record its information. In each XML file, the location of the corresponding picture and all the marked mask or nomask labels and their coordinates are recorded. In addition, all images and their corresponding XML files are divided into training part and test part, and the ratio of training and testing datasets is 9:1.

After getting the preprocessed dataset, we create a LabelMap file in pbtxt format, with a "Mask" value of 1 and a "NoMsak" value of 2. This file is used to provide auxiliary information for the subsequent marking and identification stages. Then convert all XML files and labelMaps into TFRecords files to prepare for subsequent training with TensorFlow FrameWork.

\begin{figure*}
    \centerline{\includegraphics[scale=0.5]{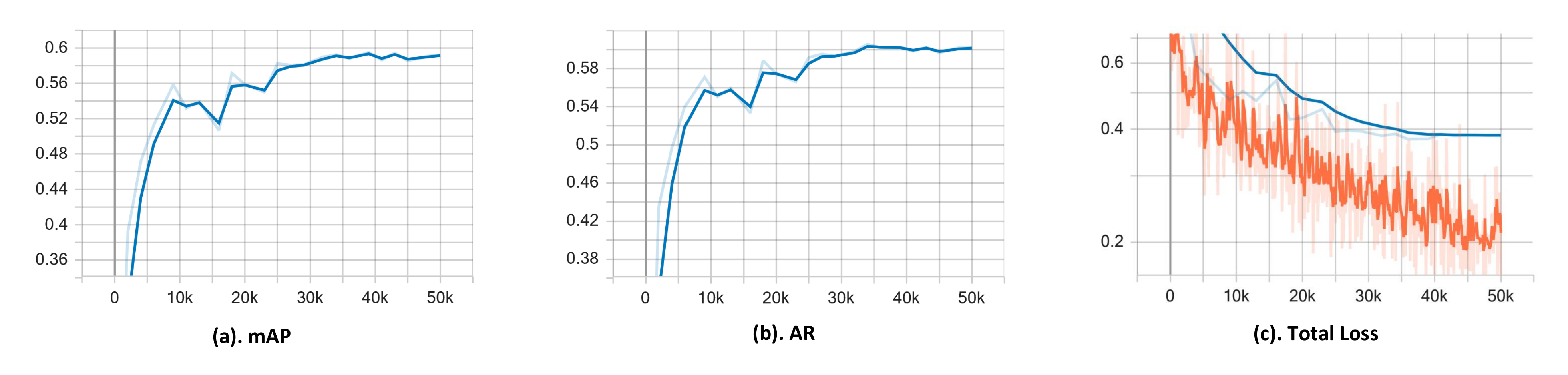}}
    \caption{model-loss}
    \label{model-loss}
\end{figure*}

\subsection{Model Training and Platform Development}
Our experiments are based on the TensorFlow Object Detection API. This framework can detect specific objects in images based on a given model, while users do not need to pay too much attention to internal details. We just need to write the $pipeline.config$ file as needed. This file contains all the configuration needed for training with the TensorFlow Object Detection API.

Since our system only recognizes whether the target is wearing a mask, $num\_classes$ is set to 2, corresponding to the labels of "Mask" and "NoMask". The transfer learning method can build our model faster, so we specify to use the pre-trained model checkpoint0 as the training starting point instead of starting from scratch. The package pyCOCOtools (python API tools of COCO) is specified as an evaluation model to evaluate our model during training. After editing all other configuration items, we use the training file provided by the TensorFlow Object Detection API and specify the previously written configuration file to start training. Each stage in the training process is saved in a folder in the form of checkpoints to facilitate building the system later.

After the model training is complete, we can start building the real-time detection system. The construction of the real-time detection system mainly relies on the OpenCV library. We load a checkpoint from the model that we trained earlier. And use OpenCv to open the local camera, get the video stream, and capture frames from the video stream. Then we can use the loaded model to perform the following processing on the video frame: First, preprocess. Crop the image size to 320*320; then put it into the model for detection, and get the returned label and location; finally, postprocess. Draw a rectangular frame on the detected target area, and convert the detected labels into the text to mark around the rectangular frame. In this way, a real-time mask detection system is constructed using our trained model.

\subsection{Performance Evaluation}
MS COCO (Microsoft Common Objects in Context) is a large-scale image dataset, usually used for object detection, segmentation, human keypoint detection, material segmentation, etc. COCO also provides a series of evaluation indicators, which is one of the most commonly used benchmarks for target detection. COCO provides more than ten indicators including average precision. We will use the COCO metric to evaluate our model. The metrics used in our model such as accuracy, precision, and recall are all based on True Positive (TP), False Positive (FP), True Negative (TN), and False Negative (FN). As shown in Equation(\ref{Accuracy}. \ref{Precision}. \ref{Recall}). From Figure \ref{model-loss} we can clearly see that the mAP, AR, and  Total Loss, denote that our model can get outstanding performance in mask detection tasks.

\begin{equation}
    \label{Accuracy}
    Accuracy = \frac{T_p + T_n}{T_p + F_p + F_n + T_n}
\end{equation}

\begin{equation}
    \label{Precision}
    Precision = \frac{T_p }{T_p + F_n}
\end{equation}

\begin{equation}
    \label{Recall}
    Recall = \frac{T_p }{T_p + F_n}
\end{equation}

TensorBoard is a tool that can visualize the parameter changes during the training process. We use tensorboard to visualize the parameters during the model training process.

\begin{figure}[H]
    \centerline{\includegraphics[scale=0.32]{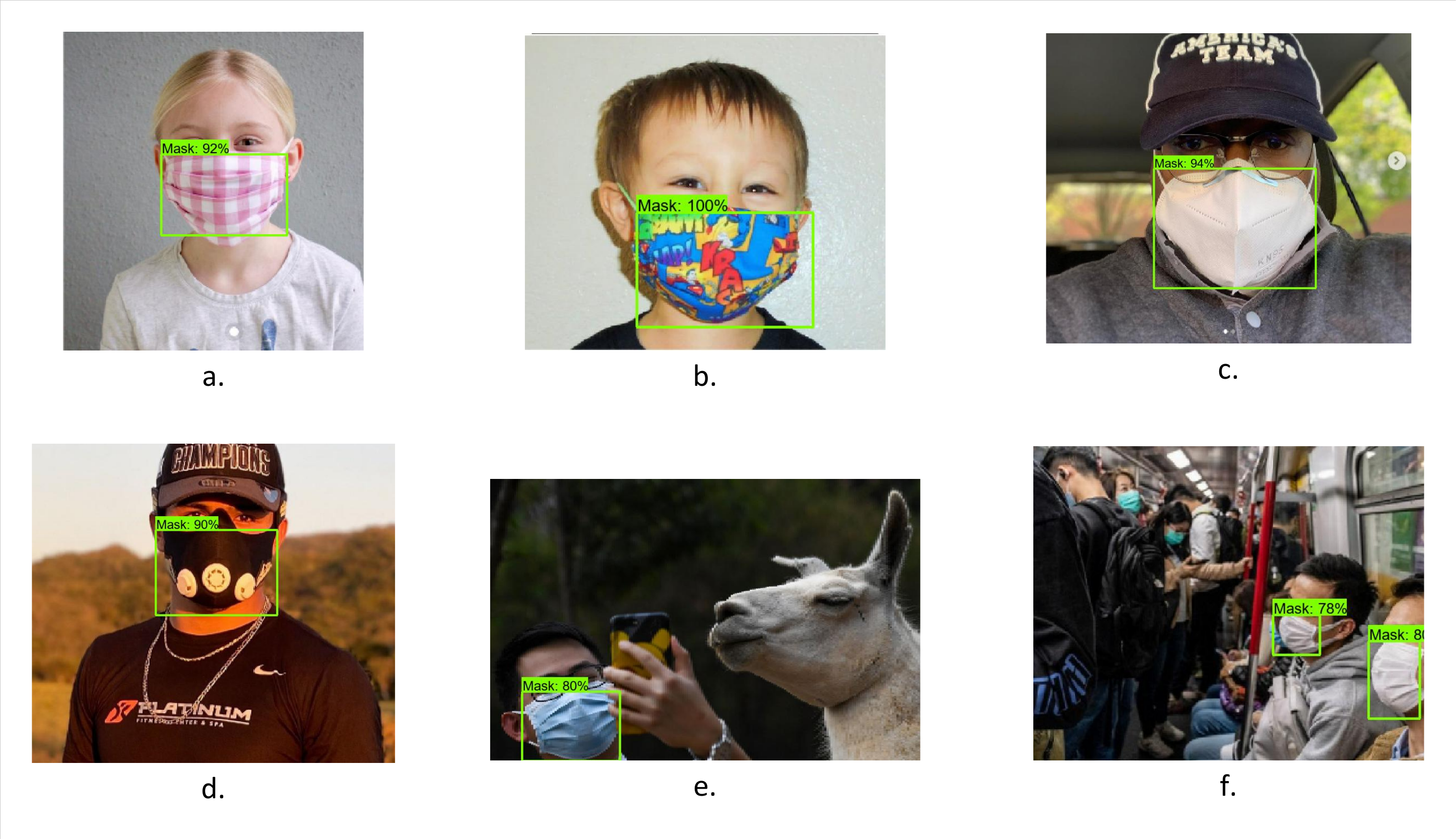}}
    \caption{Image evaluation results}
    \label{evaluation-images6}
\end{figure}

As shown in Figure \ref{evaluation-images6} (a-e), under various conditions such as dark light, occlusion, and inadequate mask-wearing, our real-time mask detection system can still accurate positioning and achieve high accuracy.

However, for a few cases with severe occlusion and strange shapes of masks, there is still the possibility of misjudgment and omission.

Real-time detection faces challenges such as random occlusion, changing features, and subject blur. Therefore, the accuracy obtained in the actual real-time detection system is lower than that obtained in the model testing process. As shown in Figure  \ref{evaluation-images6} (f)

After a large number of tests in different scenarios, our real-time mask detection system can perform well in various environments. This has great significance for practical use. Our system takes up little resources and can run on embedded devices (such as Raspberry Pi). After actual use experience, our mask detection system can make a judgment in real-time which is so short that it can meet the detection needs of the crowd in a state of continuous movement.

\section{Conclusion}
In this paper, we propose a real-time mask detection platform based on SSD-MobileNetV2 which could be helpful for preventing COVID-19. The platform can detect the status of wearing masks of people with high accuracy and fast speed. Since we replaced the VGG-16 with MobileNetV2, our model is lightweight and could be easily deployed on embedded devices. Benefiting from the data augmentation method MixUp and transfer learning, our model can be quickly trained on a small dataset. In the future, our system could be installed in stations, hospitals, schools, and other places with high population density and high traffic flow to improve the efficiency of mask detection.

\bibliographystyle{ieeetr} %ieeetr国际电气电子工程师协会期刊
\bibliography{ref.bib}

\begin{thebibliography}{10}

\bibitem{howard2020face}
J.~Howard, A.~Huang, Z.~Li, Z.~Tufekci, V.~Zdimal, H.-M. van~der Westhuizen,
  A.~von Delft, A.~Price, L.~Fridman, L.-H. Tang, {\em et~al.}, ``Face masks
  against covid-19: an evidence review,'' 2020.

\bibitem{galbadage2020does}
T.~Galbadage, B.~M. Peterson, and R.~S. Gunasekera, ``Does covid-19 spread
  through droplets alone?,'' {\em Frontiers in public health}, vol.~8, p.~163,
  2020.

\bibitem{eikenberry2020mask}
S.~E. Eikenberry, M.~Mancuso, E.~Iboi, T.~Phan, K.~Eikenberry, Y.~Kuang,
  E.~Kostelich, and A.~B. Gumel, ``To mask or not to mask: Modeling the
  potential for face mask use by the general public to curtail the covid-19
  pandemic,'' {\em Infectious disease modelling}, vol.~5, pp.~293--308, 2020.

\bibitem{verma2020visualizing}
S.~Verma, M.~Dhanak, and J.~Frankenfield, ``Visualizing the effectiveness of
  face masks in obstructing respiratory jets,'' {\em Physics of Fluids},
  vol.~32, no.~6, p.~061708, 2020.

\bibitem{qin2020identifying}
B.~Qin and D.~Li, ``Identifying facemask-wearing condition using image
  super-resolution with classification network to prevent covid-19,'' {\em
  Sensors}, vol.~20, no.~18, p.~5236, 2020.

\bibitem{jiang2021real}
X.~Jiang, T.~Gao, Z.~Zhu, and Y.~Zhao, ``Real-time face mask detection method
  based on yolov3,'' {\em Electronics}, vol.~10, no.~7, p.~837, 2021.

\bibitem{xiao2020review}
Y.~Xiao, Z.~Tian, J.~Yu, Y.~Zhang, S.~Liu, S.~Du, and X.~Lan, ``A review of
  object detection based on deep learning,'' {\em Multimedia Tools and
  Applications}, vol.~79, no.~33, pp.~23729--23791, 2020.

\bibitem{tan2020efficientdet}
M.~Tan, R.~Pang, and Q.~V. Le, ``Efficientdet: Scalable and efficient object
  detection,'' in {\em Proceedings of the IEEE/CVF conference on computer
  vision and pattern recognition}, pp.~10781--10790, 2020.

\bibitem{kong2020foveabox}
T.~Kong, F.~Sun, H.~Liu, Y.~Jiang, L.~Li, and J.~Shi, ``Foveabox: Beyound
  anchor-based object detection,'' {\em IEEE Transactions on Image Processing},
  vol.~29, pp.~7389--7398, 2020.

\bibitem{girshick2015fast}
R.~Girshick, ``Fast r-cnn,'' in {\em Proceedings of the IEEE international
  conference on computer vision}, pp.~1440--1448, 2015.

\bibitem{ren2015faster}
S.~Ren, K.~He, R.~Girshick, and J.~Sun, ``Faster r-cnn: Towards real-time
  object detection with region proposal networks,'' {\em Advances in neural
  information processing systems}, vol.~28, 2015.

\bibitem{redmon2016you}
J.~Redmon, S.~Divvala, R.~Girshick, and A.~Farhadi, ``You only look once:
  Unified, real-time object detection,'' in {\em Proceedings of the IEEE
  conference on computer vision and pattern recognition}, pp.~779--788, 2016.

\bibitem{liu2016ssd}
W.~Liu, D.~Anguelov, D.~Erhan, C.~Szegedy, S.~Reed, C.-Y. Fu, and A.~C. Berg,
  ``Ssd: Single shot multibox detector,'' in {\em European conference on
  computer vision}, pp.~21--37, Springer, 2016.

\bibitem{howard2017mobilenets}
A.~G. Howard, M.~Zhu, B.~Chen, D.~Kalenichenko, W.~Wang, T.~Weyand,
  M.~Andreetto, and H.~Adam, ``Mobilenets: Efficient convolutional neural
  networks for mobile vision applications,'' {\em arXiv preprint
  arXiv:1704.04861}, 2017.

\bibitem{huang2019rapid}
R.~Huang, J.~Gu, X.~Sun, Y.~Hou, and S.~Uddin, ``A rapid recognition method for
  electronic components based on the improved yolo-v3 network,'' {\em
  Electronics}, vol.~8, no.~8, p.~825, 2019.

\bibitem{zhao2017pedestrian}
Z.-Q. Zhao, H.~Bian, D.~Hu, W.~Cheng, and H.~Glotin, ``Pedestrian detection
  based on fast r-cnn and batch normalization,'' in {\em International
  Conference on Intelligent Computing}, pp.~735--746, Springer, 2017.

\bibitem{dai2019hybridnet}
X.~Dai, ``Hybridnet: A fast vehicle detection system for autonomous driving,''
  {\em Signal Processing: Image Communication}, vol.~70, pp.~79--88, 2019.

\bibitem{jaeger2020retina}
P.~F. Jaeger, S.~A. Kohl, S.~Bickelhaupt, F.~Isensee, T.~A. Kuder, H.-P.
  Schlemmer, and K.~H. Maier-Hein, ``Retina u-net: Embarrassingly simple
  exploitation of segmentation supervision for medical object detection,'' in
  {\em Machine Learning for Health Workshop}, pp.~171--183, PMLR, 2020.

\bibitem{mhalla2018embedded}
A.~Mhalla, T.~Chateau, S.~Gazzah, and N.~E.~B. Amara, ``An embedded
  computer-vision system for multi-object detection in traffic surveillance,''
  {\em IEEE Transactions on Intelligent Transportation Systems}, vol.~20,
  no.~11, pp.~4006--4018, 2018.

\bibitem{loey2021hybrid}
M.~Loey, G.~Manogaran, M.~H.~N. Taha, and N.~E.~M. Khalifa, ``A hybrid deep
  transfer learning model with machine learning methods for face mask detection
  in the era of the covid-19 pandemic,'' {\em Measurement}, vol.~167,
  p.~108288, 2021.

\bibitem{nagrath2021ssdmnv2}
P.~Nagrath, R.~Jain, A.~Madan, R.~Arora, P.~Kataria, and J.~Hemanth, ``Ssdmnv2:
  A real time dnn-based face mask detection system using single shot multibox
  detector and mobilenetv2,'' {\em Sustainable cities and society}, vol.~66,
  p.~102692, 2021.

\bibitem{wang2020masked}
Z.~Wang, G.~Wang, B.~Huang, Z.~Xiong, Q.~Hong, H.~Wu, P.~Yi, K.~Jiang, N.~Wang,
  Y.~Pei, {\em et~al.}, ``Masked face recognition dataset and application,''
  {\em arXiv preprint arXiv:2003.09093}, 2020.

\bibitem{jignesh2020face}
G.~Jignesh~Chowdary, N.~S. Punn, S.~K. Sonbhadra, and S.~Agarwal, ``Face mask
  detection using transfer learning of inceptionv3,'' in {\em International
  Conference on Big Data Analytics}, pp.~81--90, Springer, 2020.

\bibitem{zhu2020application}
Z.~Zhu and Y.~Cheng, ``Application of attitude tracking algorithm for face
  recognition based on opencv in the intelligent door lock,'' {\em Computer
  Communications}, vol.~154, pp.~390--397, 2020.

\bibitem{sandler2018mobilenetv2}
M.~Sandler, A.~Howard, M.~Zhu, A.~Zhmoginov, and L.-C. Chen, ``Mobilenetv2:
  Inverted residuals and linear bottlenecks,'' in {\em Proceedings of the IEEE
  conference on computer vision and pattern recognition}, pp.~4510--4520, 2018.

\bibitem{zhang2017mixup}
H.~Zhang, M.~Cisse, Y.~N. Dauphin, and D.~Lopez-Paz, ``mixup: Beyond empirical
  risk minimization,'' {\em arXiv preprint arXiv:1710.09412}, 2017.

\bibitem{chen2018face}
Q.~Chen and L.~Sang, ``Face-mask recognition for fraud prevention using
  gaussian mixture model,'' {\em Journal of Visual Communication and Image
  Representation}, vol.~55, pp.~795--801, 2018.

\bibitem{mohamed2022face}
M.~M. Mohamed, M.~A. Nessiem, A.~Batliner, C.~Bergler, S.~Hantke, M.~Schmitt,
  A.~Baird, A.~Mallol-Ragolta, V.~Karas, S.~Amiriparian, {\em et~al.}, ``Face
  mask recognition from audio: The masc database and an overview on the mask
  challenge,'' {\em Pattern Recognition}, vol.~122, p.~108361, 2022.

\bibitem{sethi2021face}
S.~Sethi, M.~Kathuria, and T.~Kaushik, ``Face mask detection using deep
  learning: An approach to reduce risk of coronavirus spread,'' {\em Journal of
  biomedical informatics}, vol.~120, p.~103848, 2021.

\bibitem{krizhevsky2012imagenet}
A.~Krizhevsky, I.~Sutskever, and G.~E. Hinton, ``Imagenet classification with
  deep convolutional neural networks,'' {\em Advances in neural information
  processing systems}, vol.~25, 2012.

\bibitem{girshick2014rich}
R.~Girshick, J.~Donahue, T.~Darrell, and J.~Malik, ``Rich feature hierarchies
  for accurate object detection and semantic segmentation,'' in {\em
  Proceedings of the IEEE conference on computer vision and pattern
  recognition}, pp.~580--587, 2014.

\end{thebibliography}

\vspace{12pt}
\color{red}

\end{document}